%% file: main.tex
\documentclass[10pt,twocolumn,letterpaper]{article}

\usepackage[pagenumbers]{cvpr}      

\input{preamble}

\definecolor{cvprblue}{rgb}{0.21,0.49,0.74}
\usepackage[pagebackref,breaklinks,colorlinks,citecolor=cvprblue]{hyperref}
\usepackage[title]{appendix}
\usepackage{subcaption}
\def\paperID{*****} 
\def\confName{CVPR}
\def\confYear{2024}

\title{Role of Uncertainty in Anticipatory Trajectory Prediction for a Ping-Pong Playing Robot
}

\author{Nima Rahmanian\\
\small University of California, Berkeley\\
{\tt\small nimsi@berkeley.edu}
\and
Michael Gupta\\
\small University of California, Berkeley\\
{\tt\small guptamvjm@berkeley.edu}
\and
Renzo Soatto\\
\small University of California, Berkeley\\
{\tt\small renzo@berkeley.edu}
\and
Srisai Nachuri\\
\small University of California, Berkeley\\
{\tt\small srisainachuri@berkeley.edu}
\and
Michael Psenka\\
\small University of California, Berkeley\\
{\tt\small psenka@eecs.berkeley.edu}
\and
Yi Ma\\
\small University of California, Berkeley\\
{\tt\small yima@eecs.berkeley.edu}
\and
S. Shankar Sastry\\
\small University of California, Berkeley\\
{\tt\small sastry@coe.berkeley.edu}
}

\usepackage{amsmath}
\usepackage{amssymb}
\usepackage{gensymb}

\begin{document}

\maketitle

\begin{abstract}
Robotic interaction in fast-paced environments presents a substantial challenge, particularly in tasks requiring the prediction of dynamic, non-stationary objects for timely and accurate responses. An example of such a task is ping-pong, where the physical limitations of a robot may prevent it from reaching its goal in the time it takes the ball to cross the table. The scene of a ping-pong match contains rich visual information of a player's movement that can allow future game state prediction, with varying degrees of uncertainty. To this aim, we present a visual modeling, prediction, and control system to inform a ping-pong playing robot utilizing visual model uncertainty to allow earlier motion of the robot throughout the game. We present demonstrations and metrics in simulation to show the benefit of incorporating model uncertainty, the limitations of current standard model uncertainty estimators, and the need for more verifiable model uncertainty estimation. Our code is publicly available.
\end{abstract}

\section{Introduction}

Ping-pong robotics offers a rich testbed for algorithmic strategies in highly dynamical games with competitive adversaries. Ping-pong has been a challenging domain of AI and robotics  for almost half a century \cite{anderson1988robot}, as we still do not have robots able to beat ping-pong masters like we have for board games such as Chess or Go \cite{silver2017mastering}. There are many visual aspects of the ping-pong setting -- an indoor setting played by two players who occupy non-overlapping regions of space with sharp colors contrasting different objects -- which allows for easier visual analysis. However, visual modeling and control remains a difficult challenge due to the highly dynamic game state and fine visual details that determine the outcome of a player's hit. As a result, many of recent advancements in ping-pong robotics have been reinforcement or deep learning focused \citep{huang2015learning,ruggiero2018nonprehensile,gao2020robotic, ding2022goalseye, wu2020futurepong}. These techniques enable robots to improve their skills through practice, much like a human player. However, despite these technological strides, ping-pong robots still face notable limitations, particularly when playing against human opponents who employ diverse and adversarial strategies. Winning a point in ping-pong is not just about reacting to the ball; it involves strategic planning throughout the full game, adapting to the opponent's play style, and exploiting their weaknesses. While work has been done on anticipatory planning before the robot's opponent hits the ball \citep{wang2017anticipatory}, such work and frameworks currently do not return confidence measures to inform a controller how much it should rely on a prediction.

Uncertainty estimation of neural models is a widely studied area \citep{gawlikowski2023survey}, and reliable uncertainty estimation is a cornerstone of neuro-symbolic computing and interpretable deep learning \citep{seshia2022toward, torfah2023learning}. Uncertainty estimation is practically crucial for advancing high-speed robotics, where computations and predictions must be made as early and frequently as possible in order to give a robotic controller enough time to reach the intended target. Scaling controller speed based on predicted uncertainty of the visual model then gives a concrete, measurable way to determine the effectiveness of an uncertainty estimation method. Further, we see from results presented in this paper that \emph{reliable} model uncertainty estimation is crucial to see true performance improvement, as many standard model uncertainty estimators over-estimate uncertainty and thus over-restrict controller movement. Using ping-pong controller performance as a concrete measure for model uncertainty estimation performance, we demonstrate that common state of the art uncertainty quantification methods fail to capture the true uncertainties in ping-pong ball trajectory prediction.

We begin by highlighting related work, both in ping-pong robotics and uncertainty estimation of deep neural networks. We then introduce our perception system and testing framework in a ping-pong playing environment, and present results and ablations of the performance of various uncertainty estimators in this environment.

\subsection{Related work}
\label{sec:related}
We now give an overview of previous work in ping-pong robotics and model uncertainty estimation. While ping-pong robotics has a long history \cite{anderson1988robot}, there has been a more recent surge in data-driven methods for controlling a ping-pong robot \cite{huang2015learning,ruggiero2018nonprehensile,gao2020robotic, ding2022goalseye}. Below, we cover explored methodologies in recent years, which emphasize data-driven methods.

\noindent \textbf{End to end learning.}
A number of works have implemented a model-free reinforcement learning approach to ping-pong robotics, using very little domain knowledge of the game of ping-pong itself \citep{gao2020robotic, buchler2022learning, zhu2018towards, ding2022goalseye}. While such approaches have seen much modern success, they typically incur a trade-off in data and sample efficiency.

\noindent \textbf{Trajectory prediction.}
A common use of domain information for ping-pong is through trajectory prediction \cite{wu2020futurepong, wang2017anticipatory}. Even after the opponent hits the ball, ping-pong ball trajectories have irregularities due to spin, the Magnus effect, and air friction, requiring more sophisticated methods than a simple quadratic extrapolation. Thus, data-driven models have been used to fit such trajectories \cite{lin2020ball}. However, the current body of work is restricted to specific settings, and uncertainties in prediction are not integrated into a control scheme.

\noindent \textbf{Uncertainty estimation} (or \emph{uncertainty quantification}). Uncertainty estimation of neural network outputs has been a long-standing problem spanning various domains \citep{gawlikowski2023survey, abdar2021review, mena2021survey}. A number of challenges with uncertainty quantification, including sensitivity to domain shifts and routinely over-estimating certainty, along with several methods to overcome these challenges, are outlined in \cite{gawlikowski2023survey}. Two dominant methodologies are \emph{ensembling methods} \citep{rahaman2021uncertainty, dolezal2022uncertainty, torfah2023learning}, where a collection of models is trained and uncertainty is computed between the separate models, and \emph{conformal prediction} \citep{romano2020classification, angelopoulos2020uncertainty}, where a series of statistical tests are performed on a model as a ``black box'' (only input to output behavior is analyzed). However, uncertainty quantification is relatively unexplored in ping-pong robotics, which offers a concrete, measurable environment to develop, test, and utilize various uncertainty quantification methods.

\section{Data-Driven anticipatory prediction}

\begin{figure}
  \centering
  \includegraphics[width=0.5\textwidth]{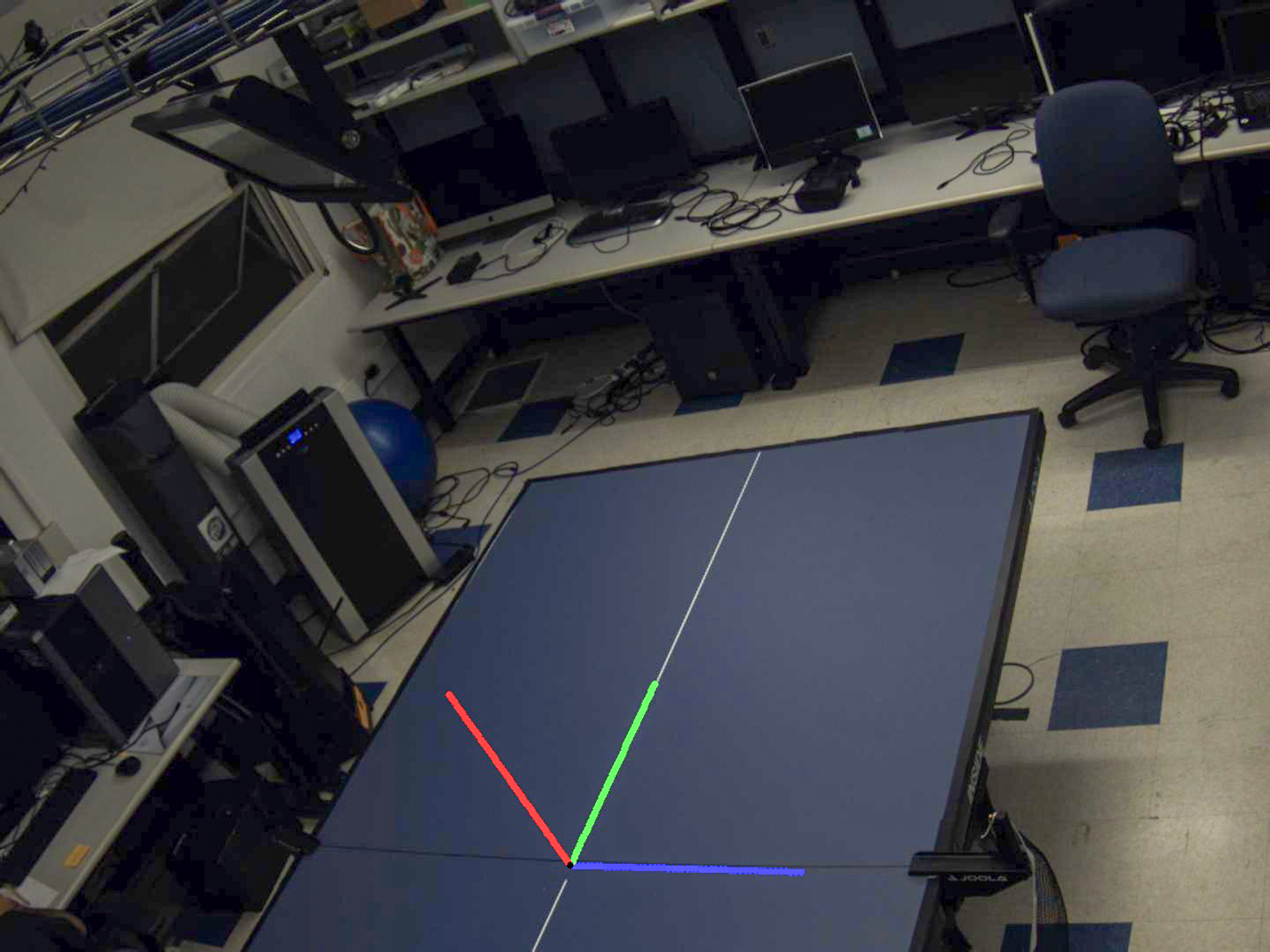}
  \caption{Spatial frame defined at the table center.}
  \label{fig:spatial_frame}
\end{figure}

We now describe the framework and evaluation procedure for our uncertainty-aware ping-pong controller. For all later discussions, we adopt the following naming conventions: the opponent is centered on the positive $y$-axis side of the table, the \emph{pre-hit} and \emph{post-hit} time intervals refer to the time before and after the opponent hits the ball. The post-hit time interval is further split into the \emph{pre-bounce} and \emph{post-bounce} time intervals, and the \emph{strike point} $s$ is defined as the $x$ and $z$ coordinates of the ball when it crosses the back of the table at $y = -140$ cm. Our ping-pong table has dimensions (152.5 x 274) cm, and we center the spatial frame on the table as shown in Figure \ref{fig:spatial_frame} with the $x$-axis along the width of the table (blue), the $y$-axis along the length (green).

The framework uses significant prior information of ping-pong, as we describe further below, in order to allow meaningful and more interpretable results in a lower data regime.
\subsection{Problem formulation and Background}
\label{method}

\begin{figure}[htp]
    \centering
    \includegraphics[width=1\linewidth]{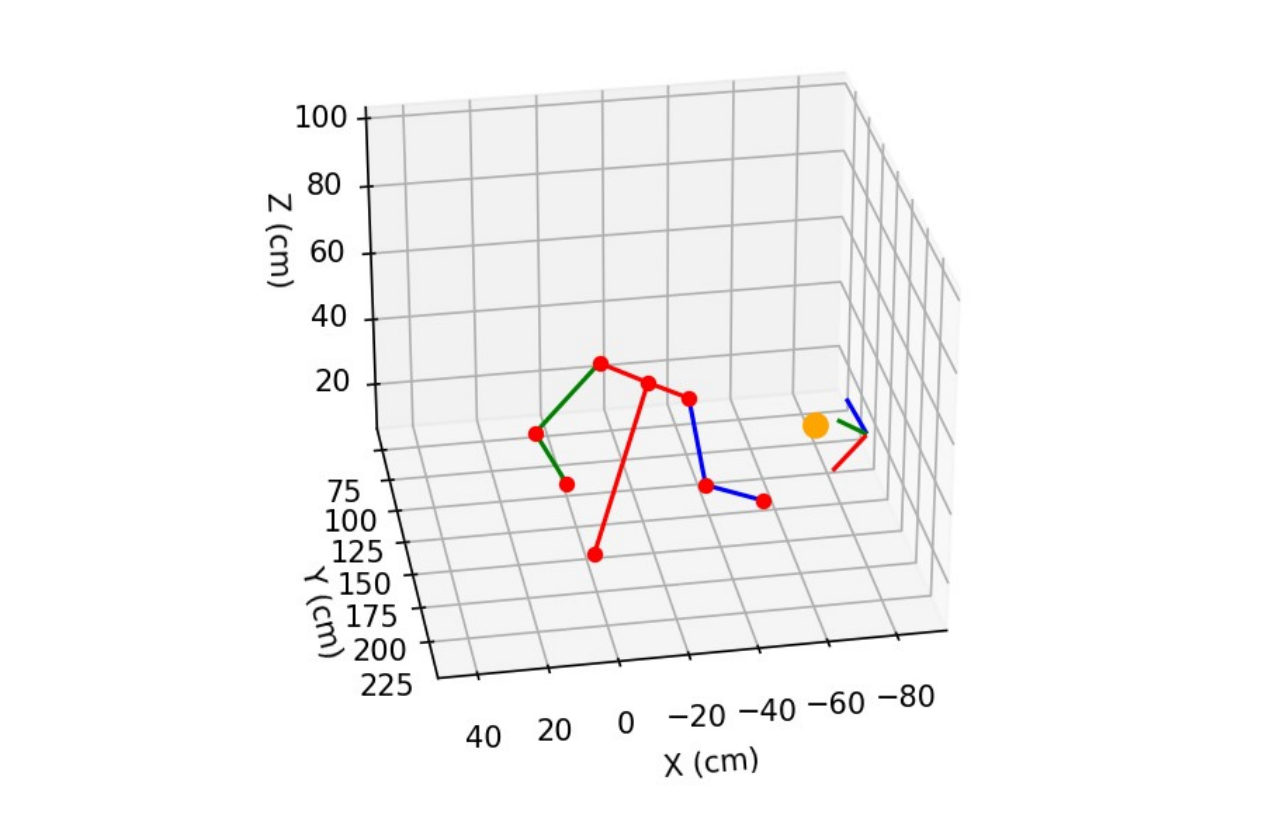}
    \caption{3D output of pose and ball triangulation.}
    \label{fig:pose_3d}
\end{figure}

We cast the anticipatory prediction problem as a classical supervised learning task. We predict the ball's post-hit trajectory using a time series of states leading up to the hit, which includes the concatenation of $1)$ the opponent's $3$-dimensional, $8$-joint human pose skeleton, $2)$ the paddle's pose in $SE(3)$, and $3)$ the ball's $3$-dimensional position, all stacked in a $\mathbb{R}^{39}$ vector. This is visualized in Figure \ref{fig:pose_3d}. We make the first anticipatory prediction $100$ms before the ball is hit, and we continue to update predictions until the ball is hit by the player. Our model learns to anticipate the player's intent and inform a controller sufficiently before the ball it hit to enhance the robot's game playing capabilities. A reliable anticipatory post-hit trajectory estimate enables the robot to build momentum towards the strike point and later hands controls off to a downstream algorithm, such as a visual servoing algorithm, to steer the end effector after the ball is hit using trajectory updates. 

\subsection{Anticipatory model objective} \label{ssec:AMO}

Any given call of the anticipatory model takes a time series of $L$ states $\mathbf{X} \in \mathbb{R}^{L \times 39}$ as input.

Since we need to pass the final strike point to the controller, the obvious solution is to predict $s$ from $\mathbf{X}$. However, this is a difficult task because of the long prediction horizon: given a set of states, the model needs to predict the ball position for a state many frames in the future. In order to reduce the horizon, we instead predict a parametrization of the ball's post-hit trajectory, starting \emph{immediately} after the contact. We observed that modeling the post-hit trajectory as two distinct, connected curves better captures complex post-bounce ball trajectories which arise due to spin. To simplify our anticipatory prediction problem, we elect not to model $z$ behavior, instead focusing on the ball's piecewise linear trajectory on the $xy$ plane: since the space of strike points vary far more in the $x$ direction than the $z$ direction, manipulator arms are better off prioritizing strike point extremities on the $x$-axis rather than $z$-axis. 

As such, the ball's path over the $x$ axis as a function of $y$ can be parametrized with $3$ parameters: pre-bounce and post-bounce slopes $a_1$ and $a_2$ and an intercept term $b$, as follows:

\begin{align}
    x_{\text{pre}}(y) &= a_1 y + b  \text{ for } y >= 0, \\ 
    x_{\text{post}}(y) &= a_2 y + b \text{ for } y < 0.
\end{align}

We train several recurrent neural network structures on the following loss:
\begin{align}
\begin{split}
    L(a_1, a_2, b; w) &= \sum_{y \in \{140,\ldots, -10\}} |x_{pre}(y) - \hat{x}_{pre}(y; w)| \\
    &+ \sum_{y \in \{-70,\ldots, -140\}} |x_{post}(y) - \hat{x}_{post}(y; w)|,
\end{split}
\end{align}
where the parameters of $\hat{x}_{pre}$ and $\hat{x}_{post}$ are jointly learned by a recurrent neural network with weights $w$. The pre-bounce term  helps with the reduction of the prediction horizon, while the post-bounce term encourages our model to make better longer-horizon predictions. The ranges were chosen by observing that most bounces occur for $y \in [-70, -10]$. This loss term offers a simple discretized estimate of the total area between the ground truth and estimated linear trajectories on the $xy$ plane.

\subsection{Model outputs with uncertainty for control}
Though incorporating anticipatory prediction alone to controllers is beneficial in high speed robotics, both demonstrated in previous work \cite{wang2017anticipatory} and Table \ref{tab:anticipatory_table}, past work notably lacks uncertainty quantification associated with the anticipatory prediction, an addition that improves both interpretability and performance. 

Exploration of uncertainty/confidence quantification is still an open problem, with many proposed solutions that work to varying degrees. To maximize compatibility, we utilize confidence (higher is better) proxies associated with anticipatory model predictions by enforcing that, when travelling towards the predicted strike point, the arm moves at a maximum velocity proportional to the associated confidence.

In an attempt to determine confidences per model output, we explored two of the most prominent techniques used for uncertainty quantification: observing the variance of an ensemble of models, and computing confidence intervals using conformal prediction. As a baseline, we trained a k-NN \citep{cover1967nearest} model to predict mean strike point errors as a function of the inputs.

We also introduce a simple, time-of-inference based confidence measure. Based on the hypothesis that predicting results earlier in a longer time horizon results in less confident predictions, combined with empirical observations that error scales linearly with time-to-hit, we assign a confidence inversely proportional to the time-to-hit. 

\subsection{Simulation for evaluation}

Standard evaluation metrics seen in other literature \citep{lin2020ball} that evaluate the mean squared trajectory prediction error, though useful, present an incomplete picture of the effectiveness of our anticipatory prediction model, which is designed to enhance a robot's ability to play high-speed ping-pong. 

For example, for a ground truth strike position $x_{strike} = 25$, predicting $x_{strike} = -25$ is far worse than predicting $x_{strike} = 75$. The former case directs the robot's momentum in the wrong direction, whereas the latter case builds the robot's momentum in the correct direction. 

Instead, we introduce an end-to-end simulation with an LBR iiwa 14 R820 KUKA robot mounted on top of an omnidirectional Ridgeback. This configuration utilizes seven revolute joints from KUKA, two prismatic joints from the Ridgeback, and a paddle mounted on the KUKA's end effector. In an attempt to model real-world hardware limitations, the simulator imposes operational position, velocity, and acceleration limits for all joints as well as a 100ms communication latency with the hardware. Introducing inertia into the robot model ensures that our simulation penalizes an agent that incorrectly anticipates the initial movement direction. We restrict the end-effector's reachable workspace to the $xz$ plane centered at $y = -140$ with the paddle's normal vector pointing towards the opponent side.

Our simulation operates in the same discrete timesteps as the pre-recorded ping-pong segments (recorded at 100hz). To simplify further discussion, we define $t$ as the timestep in a single pre-recorded ping-pong segment. The opponent strikes the ball at $t=0$, negative $t$ corresponds to pre-hit timesteps, and positive $t$ corresponds to post-hit timesteps. To simulate a discretized continuous jointspace controller and to account for communication latencies, the robot is limited to updating its goal configurations once every ten timesteps (10Hz). 

In the absence of an anticipatory prediction model, the robot solely estimates the strike point using direct trajectory estimation, a post-hit ball trajectory estimation method which fits curves to observed ball positions. In this case, the robot first estimates the strike point at timestep $t=10$ and passes this information to the visual servoing controller. In contrast, an anticipatory prediction model provides the robot with strike point estimations as early as timestep $t=-10$ (before the ball is hit). We demonstrate that informing the controller with such pre-hit estimates offers significant advantages in high-speed games. In this case, once the game state reaches 10 timesteps post-hit, the controller switches to direct trajectory estimation to guide the visual servoing algorithm. Details on the control law can be found in Appendix \ref{sec:controls}.

\section{Experiments and results}
We now detail the paper's main results. We certify both that our trained model can reliably predict ball trajectories from various points back in time, with varying degrees of uncertainty, and that current model uncertainty estimators do not accurately estimate the model's true expected error.

\subsection{Experimental Setup}

\noindent \textbf{Data collection: } We have extracted over 2000 hits from four hours of ping-pong-playing footage filmed on four RGB cameras sampling at 100Hz and eight OptiTrack InfraRed (IR) Cameras sampling at or above 120Hz. Ball positions in each RGB image were determined using standard color-based segmentation (algorithm details in Appendix \ref{sec:lab-setup}), and player joint positions were determined using OpenPose \cite{osokin2018real}. Both were then embedded in 3D space using standard calibration and triangulation tools from Multiview Geometry \cite{bradski2000opencv}. The paddle pose was identified using IR balls attached to the paddles, which we found resulted in higher fidelity measurements than using planar homography with the RGB images, which was much more volatile and sensitive to motion blurring. We utilized the OptiTrack software's automatic calibration and triangulation software, both of which depend on similar Multiview Geometry machinery as our RGB camera calibration and triangulation.

We also extensively filtered and curated our hits to yield our final dataset. We remove any invalid hits, including but not limited to segments where balls don't properly land on the other side of the table (e.g. hitting the net, double bounces, etc.). To counteract the effect of noisy measurements, we also low-pass filter the ball position and human pose signals, which we found greatly improved ground truth curve fitting. It is important to mention that our dataset contains an uneven distribution of hits towards the left, center, and right sides of the table due our group's playing habits.

\noindent \textbf{Anticipatory model training and architecture: }
We train on an 80-20 train-test split. More details of our training can be found in Appendix \ref{sec:hyperparameter-sweep}. Our experimental network uses an RNN block whose outputs are fed into an MLP; The RNN hidden size is 100, we use 3 layers, and dropout probability $p_{dropout}$ = 0.15. The MLP has 1 hidden layer with size 600. The final output size of our network is $\mathbf{Y} \in \mathbb{R}^{30 \times 3}$, where the $i$-th row in $\mathbf{Y}$ represents a prediction of the aforementioned 3 trajectory parameters in Section \ref{ssec:AMO} at $i$ frames before the ball is hit. In this way, predictions can be generated for any timestep between 30 frames before the ball is hit and 0 frames before the ball is hit.  

\subsection{Uncertainty estimation}

\begin{figure}[ht!]
    \includegraphics[width=1.05\linewidth]{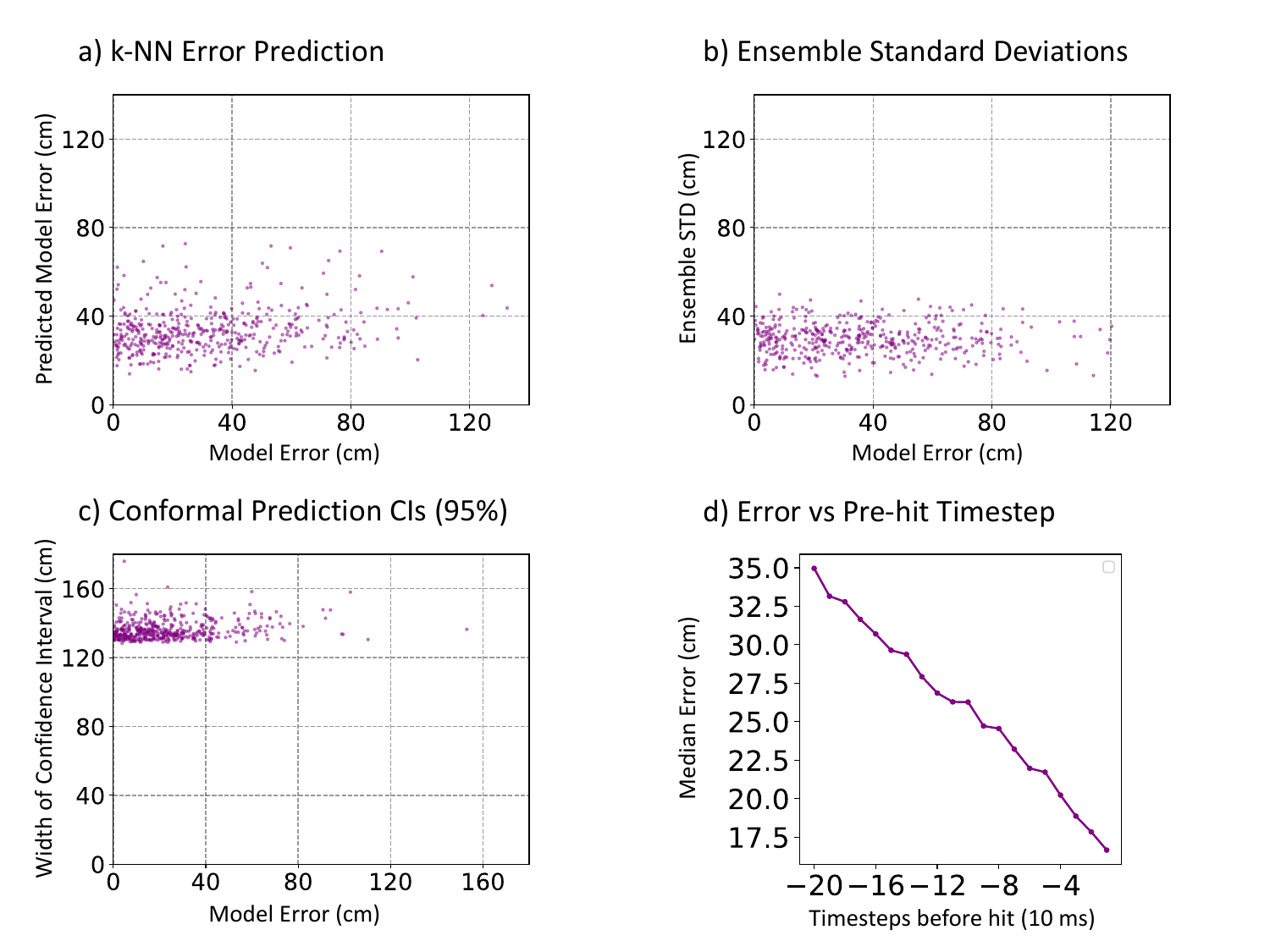}
    \caption{\textbf{a, b, c}: Samples of model error versus estimated model uncertainty for predictions made at $t=0$. For a verifiable or predictive measure of model uncertainty, there should be a strong direct correlation between the two.  \textbf{d}: Median prediction error as a function of prehit timestep. }
    \label{fig:uncertainty}
\end{figure}

Our exploration of uncertainty estimation methods indicated that accurately quantifying confidence of our anticipatory prediction method for each prediction is a very difficult task that the most common existing measures do not solve. A good confidence measure should have a strong correlation with the residual of its corresponding prediction, a trend that is notably absent across our k-NN error predictions (Fig \hyperref[fig:uncertainty]{\ref*{fig:uncertainty}a}), ensemble standard deviations (Fig \hyperref[fig:uncertainty]{\ref*{fig:uncertainty}b}), and the width of the conformal prediction confidence intervals (Fig \hyperref[fig:uncertainty]{\ref*{fig:uncertainty}c}). An interesting observation is that our k-NN error predictions and our ensemble standard deviations are clustered around $25$, which is approximately equal to the average strike point error of our model. Conformal prediction, on the other hand, is vastly over-conservative, due to its mathematical guarantees. Given our table's width is around $150$cm, confidence intervals roughly of width $130$cm are not informative.

Given that predicting uncertainty on a per-hit basis seems infeasible, another option is predicting uncertainty as a function of the time-to-hit. This is quite trivial, since the relationship between the time-to-hit and the strike point error is almost perfectly linear (Fig \hyperref[fig:uncertainty]{\ref*{fig:uncertainty}d}). As we make anticipatory predictions at various points in time, this time-to-hit is still more useful than simply quantifying the overall uncertainty of the model.

\subsection{Results on Simulation}

\begin{table}[h]
  \centering
  \begin{tabular}{|c|c|c|}
    \hline
    \multicolumn{1}{|c|}{\textbf{Frame}} & \multicolumn{1}{c|}{\textbf{10}} & \multicolumn{1}{c|}{\textbf{1}} \\
    \hline
    \textbf{Right} & 38.15 & 28.12 \\
    \hline
    \textbf{Center} & 30.62 & 18.21 \\
    \hline
    \textbf{Left} & 18.73 & 13.01 \\
    \hline
    \textbf{All} & 26.27 & 16.65 \\
    \hline
  \end{tabular}
  \caption{Median strike point error reduction over frames before the hit (in centimeters).}
  \label{tab:median_strike}
\end{table}

\begin{table}[h]
  \centering
  \begin{tabular}{|c|c|c|c|}
    \hline
    \multicolumn{1}{|c|}{} & \textbf{Total} & \textbf{\# hit} & \textbf{End dist. to goal} \\ 
    \hline
     \textbf{Right} & 101 & 60 & 10.13 $\pm$ 4.38 (cm) \\
    \hline
    \textbf{Center} & 81 & 60 & 6.26 $\pm$ 1.93 (cm) \\
    \hline
    \textbf{Left} & 41 & 24 & 12.82 $\pm$ 7.31 (cm) \\
    \hline
     \textbf{All} & 223 & 144 & 9.22 $\pm$ 4.615 (cm) \\
    \hline
  \end{tabular}
  \caption{$\alpha_1 = 0.0, \alpha_2 = 0.0$: Performance of baseline visual servoing controller. End distance to goal reported as mean $\pm$ 0.5 standard deviations. }
  \label{tab:servo_table}
\end{table}

\begin{table}[h]
  \centering
  \begin{tabular}{|c|c|c|c|}
    \hline
    \multicolumn{1}{|c|}{} & \textbf{Total} & \textbf{\# hit} & \textbf{End dist. to goal} \\ 
    \hline
     \textbf{Right} & 101 & 67 & 8.33 $\pm$ 3.85 (cm) \\
    \hline
    \textbf{Center} & 81 & 59 & 5.9 $\pm$ 1.87 (cm) \\
    \hline
    \textbf{Left} & 41 & 20 & 13.63 $\pm$ 7.41 (cm) \\
    \hline
     \textbf{All} & 223 & 146 & 8.42 $\pm$ 4.46 (cm) \\
    \hline
  \end{tabular}
  \caption{$\alpha_1 = 1.0, \alpha_2 = 1.0$: Performance of basic anticipatory controller. End distance to goal reported as mean $\pm$ 0.5 standard deviations. }
  \label{tab:anticipatory_table}
\end{table}

\begin{table}[h]
  \centering
  \begin{tabular}{|c|c|c|c|}
    \hline
    \multicolumn{1}{|c|}{} & \textbf{Total} & \textbf{\# hit} & \textbf{End dist. to goal} \\ 
    \hline
     \textbf{Right} & 101 & 70 & 8.35 $\pm$ 3.72 (cm) \\
    \hline
    \textbf{Center} & 81 & 59 & 5.84 $\pm$ 1.78 (cm) \\
    \hline
    \textbf{Left} & 41 & 22 & 12.5 $\pm$ 7.18 (cm) \\
    \hline
     \textbf{All} & 223 & 151 & 8.2 $\pm$ 4.27 (cm) \\
    \hline
  \end{tabular}
  \caption{$\alpha_1 = 0.6, \alpha_2 = 1.0$: Performance of uncertainty-aware anticipatory controller. End distance to goal reported as mean $\pm$ 0.5 standard deviations. }
  \label{tab:u_anticipatory_confidence_table}
\end{table}

While our recorded hits are sampled from both sides of the table, for simplicity, we arrange all hits such that the strike points are on the $-y$-axis side of the table. We define the ping-pong table's left, center, and right regions as all points with $x \in [-\infty, -25]$, $x \in [-25, 25]$, and $x \in [25, \infty]$, respectively. We define $\alpha_1, \alpha_2 \in [0,1]$ as the ratio of the commanded end effector velocity over the maximum allowable end effector velocity of the robot for predictions made at $t=-10$ and $t=0$, respectively. As a sanity check, we computed the RMSE in strike point (Table \ref{tab:median_strike}) and verified that predictions made closer to the time the player hits the ball have a smaller median strike point error. Though these values inform us about the model's predictive capabilities, the anticipatory planner relies most on the model's ability to guide the robot arm in the correct general direction. 

We provide two main metrics, one of which is the number of successful hits by the paddle surface at the strike point. We define \emph{end distance to goal} as the distance from the center of the paddle to ball when the ball crosses the strike point. By this definition, a successful hit may still have a nonzero \emph{end distance to goal}. Comparing the change in performance from the visual servoing baseline and the basic anticipatory controller in Table \ref{tab:servo_table} and Table \ref{tab:anticipatory_table}, we observe that while the average and standard deviation of \emph{end distance to goal} decreases for center and right hits, these metrics \emph{increase} for left hits. In these cases, we observed that the model incorrectly predicted the goal to have a positive $x$ value that lands in the center region. This suggests that when the model is uncertain about the post-hit trajectory, it predicts strike points towards the center-right table region (likely due to the right skew in the overall dataset). This in turn prompts the robot to build momentum to the \emph{right}. We also observed that the model generally predicts the correct side of the table when its strike point predictions lie in the left or right regions. With these observations in mind, we designed our uncertainty-aware controller to only make anticipatory movements when the strike point predictions lie outside of the center region. 

We determined the value of $\alpha_1$ and $\alpha_2$ by running a hyperparameter sweep over a calibration set and optimizing for number of successful hits. As seen in the performance increases from Table \ref{tab:anticipatory_table} to Table \ref{tab:u_anticipatory_confidence_table}, this uncertainty-aware design choice boosted the controller's ability to return all types of hits. While the shown improvement is marginal, the reported improvement given even such a simplistic uncertainty estimator highlights the value of reliable uncertainty estimators in their direct role on enhancing ping-pong playing ability.

\section{Discussion}
The most notable reported result is the lack of informative uncertainty prediction for standard model uncertainty estimators like variance of ensemble predictions and conformal prediction. We firstly discuss the reliability of these results, which is dependent on the trained model itself. While our model has not been engineered to a state-of-the-art performance level compared to heavily trained models on large datasets, our model fits match expected trends in uncertainty well enough (e.g. uncertainty increases as prediction inputs move farther back in time) that the lack of correlation presented in Figure \ref{fig:uncertainty} is significant.

We argue that these results call for more reliable uncertainty awareness in \emph{both directions}: the uncertainty measure should both be high enough when the model is uncertain, but low enough when the model is confident. Such accuracy, we argue, requires further interpretability and knowledge about a model's interaction with data. In exploring such interpretable uncertainty quantification measures, the relative simplicity of our model (simple RNN, using interpretable representations of the ping-pong game state) becomes an asset for reliably determining factors and potential avenues for the development of accurate and reliable uncertainty estimators.

\section{Conclusion and future work}

Ping pong robotics remains an interesting application for modern visual models, as the high speed of the game requires inferences to be made well in advance, using more general cues like human pose motion, before the state of the ball is well defined. Additionally, due to the high speed of the game, reliable uncertainty prediction becomes crucial in order to inform how slow or fast a downstream controller should move, and the performance of this controller becomes a concrete metric for evaluating the reliability of the model uncertainty estimator. In the particular case of modeling the trajectory of the ping-pong ball using human pose motion, we find that standard model uncertainty quantification methods fail to reliably predict the true uncertainty of the prediction model, rendering them unhelpful to a robotic controller.

We believe that performance metrics for high-speed uncertain robotics applications, such as ping-pong, offer an interesting benchmark for future work in model uncertainty quantification for visual models. As the game has been heavily studied both in the context of classical fully model-based robotics and modern model-free reinforcement learning, ping-pong proves to be an excellent environment to explore model uncertainty quantification in a more interpretable manner, hopefully leading to more reliable and accurate model uncertainty estimators for more general applications.

{
    \small
    \bibliographystyle{ieeenat_fullname}
    \bibliography{main}
}
\newpage
\newpage

\begin{appendices}

\section{Lab Setup}
\label{sec:lab-setup}

\subsection{Hardware}

\begin{figure}[htp]
    \centering
    \includegraphics[width=0.5\linewidth]{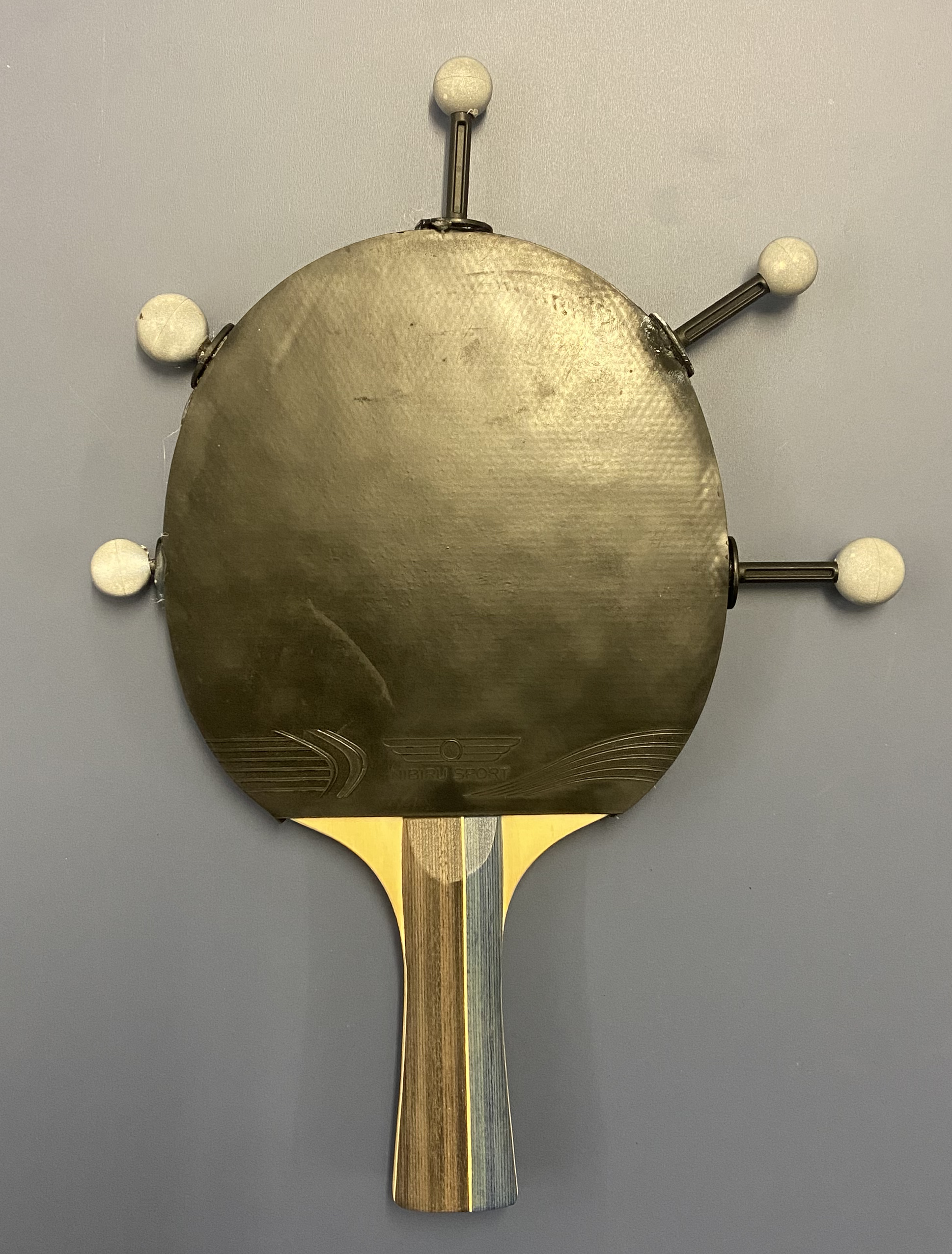}
    \caption{Infrared marker placement on ping-pong paddle.}
    \label{fig:paddle_ir_balls}
\end{figure}

Our sensor suite includes four RGB Daheng Imaging MER2 - 160-227U3C cameras (100hz) and eight OptiTrack cameras coordinated with motion capture software (at or above 120hz). We use green screen backdrops to remove orange-like objects in the background from the RGB video feed. Figure \ref{fig:paddle_ir_balls} illustrates the infrared marker placement, used by the OptiTrack motion capture system, to capture the poses of the two ping-pong paddles.

\subsection{Triangulation}

3D locations for both ball and joint locations are found using a standard direct linear transformation triangulation technique. For each camera correspondence (at least two needed), we set up the equation $U = \alpha P X$ where $U$ is the 2D pixel coordinate, $\alpha$ is a normalizing constant for defining the world scale, $P$ is the camera's corresponding projection matrix, and $X$ is the unknown 3D spatial coordinate. We stack the equations and solve for $X$ using the Direct Linear Transformation.

\subsection{Ball Detection}
\begin{figure}[htp]
    \centering
    \includegraphics[width=0.6\linewidth]{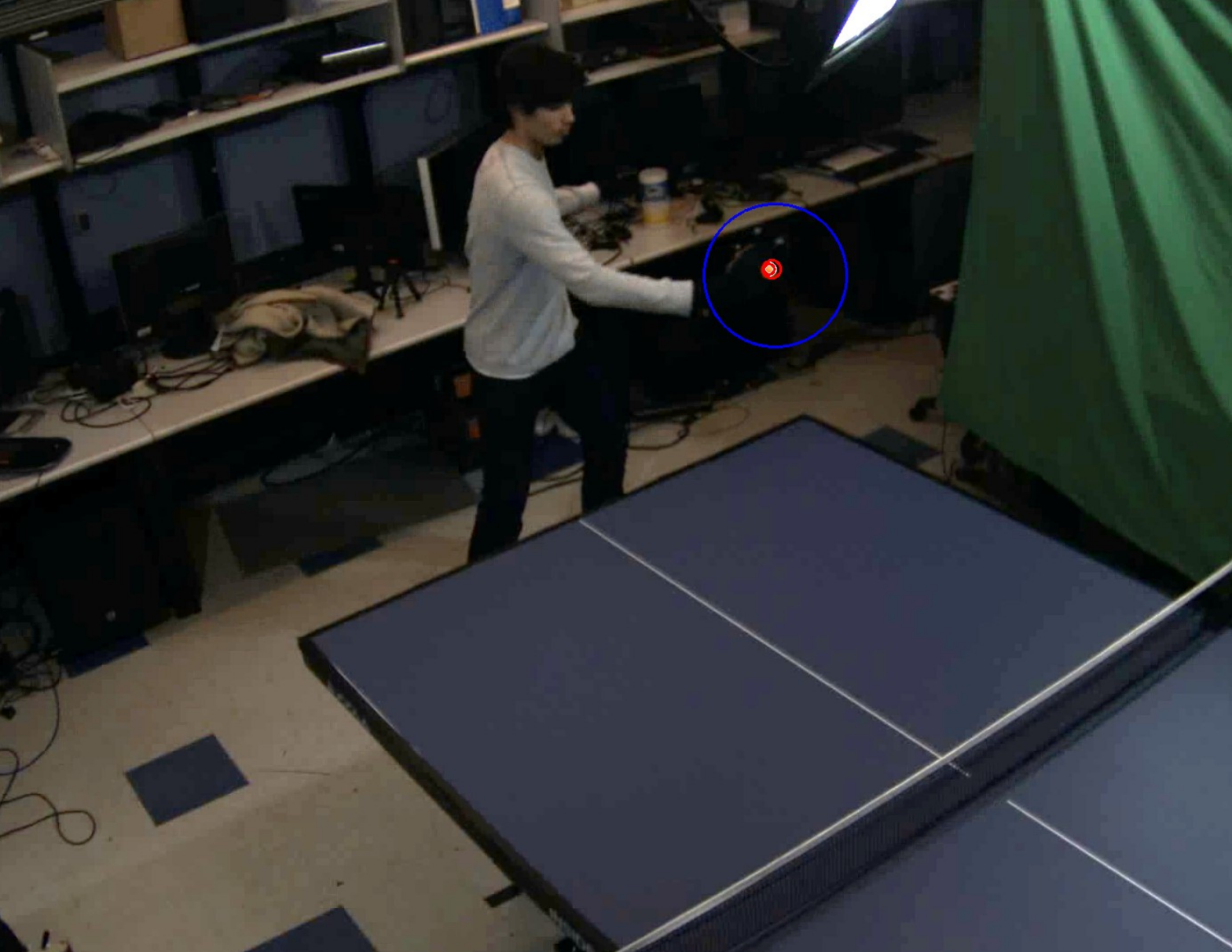}
    \caption{Isolation of ping-pong ball in 2D using color filtering. The blue circle represents the search area based  on the last ball position, and the red circle represents the contour found through masking.}
    \label{fig:ball_detect}
\end{figure}

\begin{figure}[htp]
    \centering
    \includegraphics[width=1\linewidth]{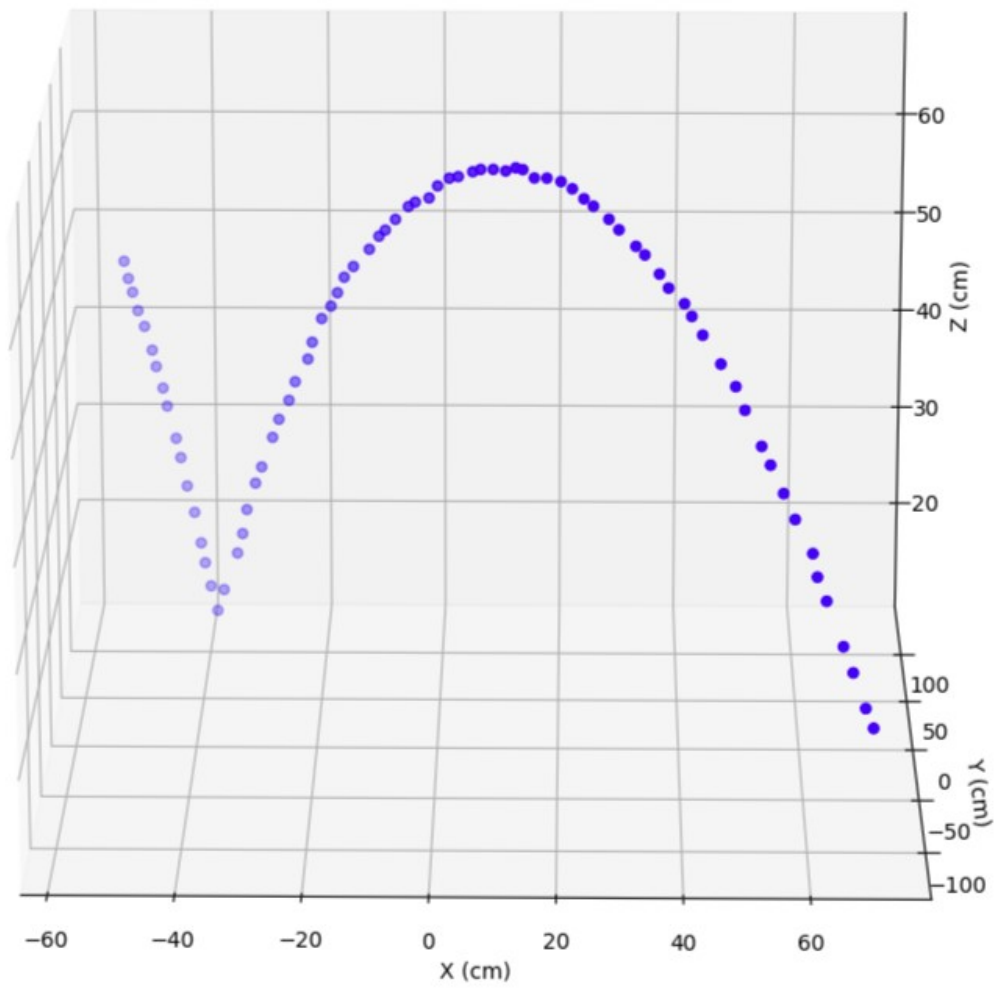}
    \caption{Example trajectory found through color filtering and triangulation. The ball is going away from the perspective of the viewer.}
    \label{fig:ex_trajectory}
\end{figure}

Ball detection from video frames is achieved through a series of masks that isolate the unique orange color of the ping-pong ball. This masking can be done especially reliably and quickly if a ball is detected in a previous frame; In this case, we only sample in the 200 pixels surrounding that previous position. With this mask, we generate contours, and find the center of the largest contour. A visualization of our ball detection technique is shown in Figure \ref{fig:ball_detect}. An example of a full trajectory found with the ball detection and triangulation systems is shown in Figure \ref{fig:ex_trajectory}.

\subsection{Human and Paddle Pose Estimation}

\begin{figure}[htp]
    \centering
    \includegraphics[width=0.6\linewidth]{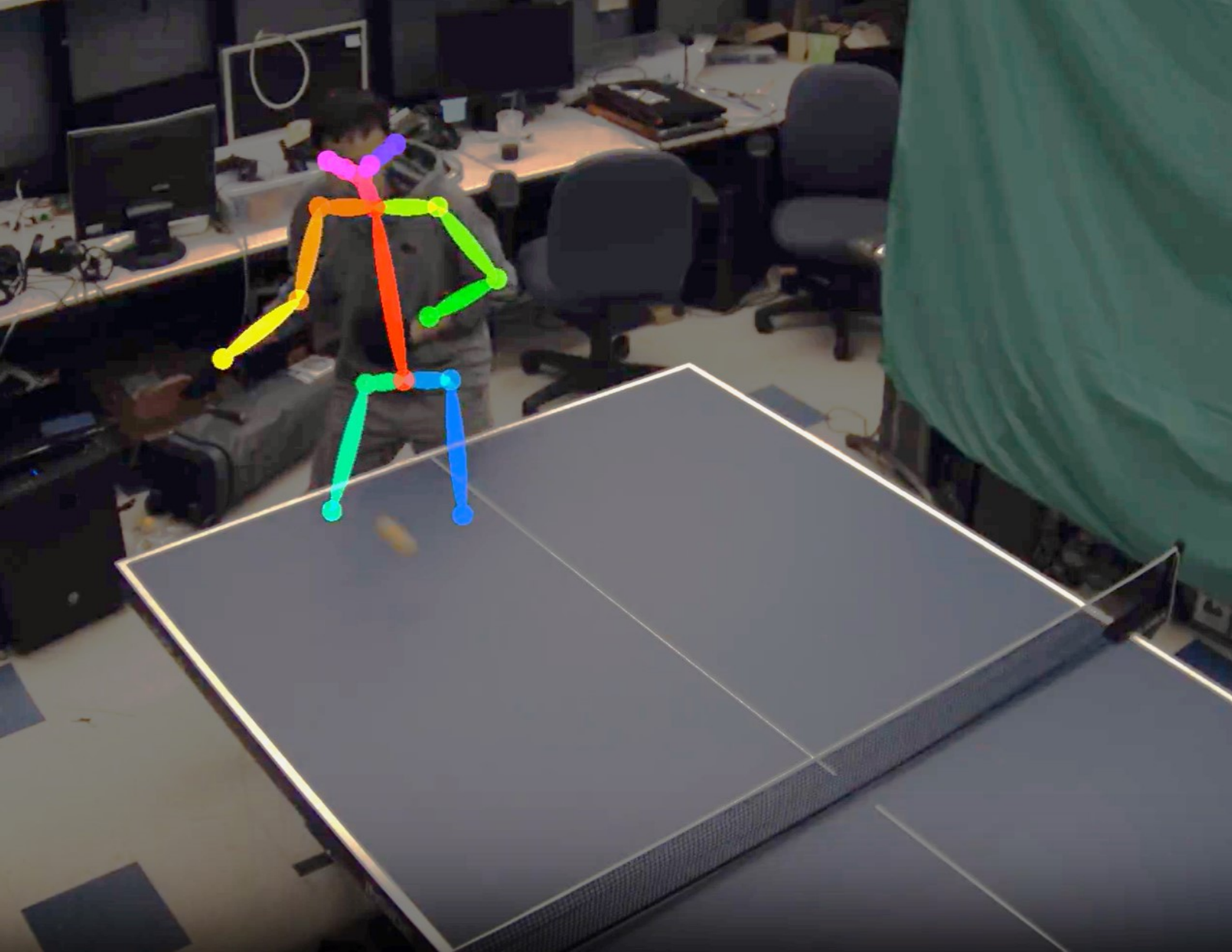}
    \caption{2D output of Openpose.}
    \label{fig:pose_2d}
\end{figure}

For pose estimation, we use OpenPose with no modifications \cite{osokin2018real}. With one Nvidia GeForce RTX 4090 GPU, it is capable of real time (60Hz) pose estimation. An example 2D pose estimation using this software is shown in Fig. \ref{fig:pose_2d}.

Given two or more views of 2D human poses and ping-pong ball positions, we calculate the 3D human poses and ball positions using the aforementioned triangulation system. The OptiTrack motion capture system provides us with the $SE(3)$ paddle poses. The 3D positions of the ball, bodies, and the paddle orientations constitute one frame of input in our dataset. A visual of this fully triangulated frame is shown in Figure \ref{fig:pose_3d}. 

\subsection{Segmentation Process }
We define a \emph{segment} as the series of frames starting when the ball crosses the net towards one player and ending when the other player strikes the ball. A segment encapsulates a player's response to the incoming ball and the associated response. We exclude segments with the following properties in our dataset:

\begin{enumerate}
    \item Sufficiently many frames with undetected or misdetected human joints in the pre-hit segment. 
    \item Sufficiently many frames with undetected paddle poses in the pre-hit segment.
    \item The post-hit ball trajectory does not bounce on the correct side of the table. 
\end{enumerate}

For segments which do enter the dataset, the ball trajectory and human pose estimate signals are low-pass filtered. We compiled a dataset of 2,226 segments.

\section{Simulation}
\label{sec:controls}
\subsection{Robot Mechanical Design}
To provide the clearest picture of the effectiveness our model, we base our simulation on the realistic setup of an LBR iiwa 14 R820 KUKA robot mounted on top of an omnidirectional Ridgeback. The KUKA, in accordance with real model specifications \footnote{\url{https://www.reeco.co.uk/wp-content/uploads/2020/05/KUKA-LBR-iiwa-technical-data.pdf}}, is represented as 7 revolute joints. The omnidirectional Ridgeback \footnote{\url{https://clearpathrobotics.com/ridgeback-indoor-robot-platform/}}, as an appropriate simplication of planar motion, is represented by 2 prismatic joints along the $xy$ plane. We enforce physical constraints provided by the robot technical specifications for joint position, velocity, and torque. 

We denote the 7 revolute joints of the KUKA ordered $A1-A7$ and the 2 prismatic joints of the Ridgeback as $X1$ and $Y1$. Range of motion (ROM), torque, and velocity constraints are provided for the KUKA while only velocity constraints are provided for the Ridgeback.

\begin{table}[h]
  \centering
  \begin{tabular}{|c|c|c|c|}
    \hline
    \multicolumn{1}{|c|}{\textbf{Axis}} & \multicolumn{1}{c|}{\textbf{ROM}} & \multicolumn{1}{c|}{\textbf{Max Torque}} & \multicolumn{1}{c|}{\textbf{Max Velocity}} \\
    \hline
    \textbf{X1} &---  & ---  &  1.1 $m/s$ \\
    \hline
    \textbf{Y1} & ---  & --- &  1.1 $m/s$ \\
    \hline
    \textbf{A1} &  $\pm 170 \degree$ &  320 $Nm$&  $85 \degree/s$ \\
    \hline
    \textbf{A2} &  $\pm 120 \degree$ &  320 $Nm$&  $85 \degree/s$ \\
    \hline
    \textbf{A3} &  $\pm 170 \degree$ &  176 $Nm$&  $100 \degree/s$ \\
    \hline
    \textbf{A4} &  $\pm 120 \degree$ &  176 $Nm$&  $75 \degree/s$ \\
    \hline
    \textbf{A5} &  $\pm 170 \degree$ &  110 $Nm$&  $130 \degree/s$ \\
    \hline
    \textbf{A6} &  $\pm 120 \degree$ &  40 $Nm$&  $135 \degree/s$ \\
    \hline
    \textbf{A7} &  $\pm 175 \degree$ &  40 $Nm$&  $135 \degree/s$ \\
    \hline
  \end{tabular}
  \caption{Provided joint limits of KUKA and Ridgeback}  
\end{table}
For workspace controllability, we impose additional constraints on the Ridgeback and extrapolate acceleration constraints for the KUKA from torque. The Ridgeback is constrainted to move along the plane within a square of area $4\text{ }m^2$, and capped at a maximum acceleration of $15\text{ }m/s^{2}$. This acceleration was chosen empirically through documentation of other similar planar robots.

KUKA acceleration constraints are derived from torque constraints by considering the mass distribution of the 8 effective links separated by 7 joints and gravitational torque resisting desired angular acceleration direction. Since both link mass distribution and gravitational torque vary based on the instantaneous configurations of the joints, we limit acceleration by the worst case scenario, in which mass is distributed furthest from the axis of rotation and the gravitational torque opposing desired angular acceleration is of its highest magnitude. For instance, a joint rotating along the y-axis, would accelerate the slowest when its target links are all orientated perpendicular to the z-axis and parallel to the x-axis. This acceleration can be reasonably applied with the max torque available for all configurations, and a controller can realistically command joint velocities to obey these limits. Acceleration constraints are applied equally for all velocity changes including braking and reversing direction.

Table 6 showcases the effective ROM, acceleration, and velocity constraints of the KUKA-Ridgeback setup. Note that while effective accelerations for joints are high, low velocity limits do not allow a sustained acceleration over a long time frame.

\begin{table}[h]
  \centering
  \begin{tabular}{|c|c|c|c|}
    \hline
    \multicolumn{1}{|c|}{\textbf{Axis}} & \multicolumn{1}{c|}{\textbf{ROM}} & \multicolumn{1}{c|}{\textbf{Max Acceleration}} & \multicolumn{1}{c|}{\textbf{Max Velocity}} \\
    \hline
    \textbf{X1} & $\pm 1$ $m$  &  15 $m/s^2$ &  1.1 $m/s$ \\
    \hline
    \textbf{Y1} & $\pm 1$ $m$  & 15 $m/s^2$ &  1.1 $m/s$ \\
    \hline
    \textbf{A1} &  $\pm 170 \degree$ & $3.69\mathrm{e}{3}$ $\degree/s^2$&  $85 \degree/s$ \\
    \hline
    \textbf{A2} &  $\pm 120 \degree$ & $3.47\mathrm{e}{3}$ $\degree/s^2$&  $85 \degree/s$ \\
    \hline
    \textbf{A3} &  $\pm 170 \degree$ &  $7.42\mathrm{e}{3}$ $\degree/s^2$&  $100 \degree/s$ \\
    \hline
    \textbf{A4} &  $\pm 120 \degree$ &  $1.37\mathrm{e}{4}$ $\degree/s^2$&  $75 \degree/s$ \\
    \hline
    \textbf{A5} &  $\pm 170 \degree$ &  $3.79\mathrm{e}{4}$ $\degree/s^2$&  $130 \degree/s$ \\
    \hline
    \textbf{A6} &  $\pm 120 \degree$ &  $3.81\mathrm{e}{5}$ $\degree/s^2$&  $135 \degree/s$ \\
    \hline
    \textbf{A7} &  $\pm 175 \degree$ &  $5.72\mathrm{e}{5}$ $\degree/s^2$&  $135 \degree/s$ \\
    \hline
  \end{tabular}
  \caption{Effective joint limits of KUKA and Ridgeback}  
\end{table}

\subsection{Workspace Controller}

We model our simulated paddle as a 8 centimeter-radius circular plane positioned at the KUKA end effector, with the paddle plane facing the opponent. We control the paddle as programmed velocities and goal positions over discrete timesteps, and thus our robot controller translates the desired workspace velocity into jointspace velocities of the KUKA and Ridgeback joints with the appropriate joint constraints applied while also incorporating feedback of the error in position. 
\subsubsection{Workspace Control Law}
First, we define our controller that maps joint velocities from paddle velocity. Using the error in the positional configuration of the end effector (paddle) and the desired workspace velocity, our controller implements a closed-loop Cartesian control law in Section 1.2.3 \footnote{\url{https://ucb-ee106.github.io/106b-sp23site/assets/proj/proj1b.pdf}}. Incorporating feedback, we aim for a workspace end effector velocity of $U^s$. As specified in the aforementioned material, we use the spatial Jacobian of the current robot configuration, $J(\theta)$ to find the minimum norm solution for joint velocities, $\dot{\theta}$, through the following equation:
\begin{equation}
\dot{\theta} = J^\dag(\theta)U^s,
\end{equation}
where $J^\dag$ is the Moore-Penrose inverse of $J$. In order to account for the discretization of workspace velocity control and to approximate signal latency, we perform control steps in increments of 0.1 $s$. The joint positions are discretely updated through the following equation:
\begin{equation}
\theta_{t+1} = \theta_{t} + 0.1\times\dot{\theta_{t}}.
\end{equation}

\subsubsection{Trajectory and Enforcement of Joint Limits}

We modify our joint position update law based on two criteria. First, we
enforce acceleration, velocity, and position constraints of the robot. Second, we follow a naive linear trajectory and desire to move the robot by a proportion $\alpha$, where $0 \leq \alpha \leq 1.0$, of the maximum possible joint speed. This second criterion allows for controllability in correspondence with uncertainty.

First we enforce robot constraints. Given our control law computed $\dot{\theta}_t$, we consider previous joint velocities, maximum joint accelerations, and maximum joint speeds to find a $\beta$ where $-1 \leq \beta \leq 1$, such that $\dot{\theta}^*_t = \beta \dot{\theta}_t$ obeys velocity and acceleration constraints and scales the joint velocity to still allow the end effector to move in the correct direction at maximum possible joint speeds. To enforce positional constraints, we simply bound our computed $\theta_{t+1}$ by the maximum and minimum positions, $\theta_{max}$ and $\theta_{min}$.

Given our maximal speed joint velocities $\dot{\theta}^*_t$, we further scale this entity by $\alpha$, which is the desired proportion of the maximum joint speed we want the robot to move with. The positional update law is thus given by the following:

\begin{equation}
\theta_{t+1} = \min\{\max\{\theta_{t} + 0.1\times \alpha\beta\dot{\theta_{t}}, \theta_{min}\}, \theta_{max}\}.
\end{equation}

\begin{figure}[htp]
    \centering
    \includegraphics[width=1\linewidth]{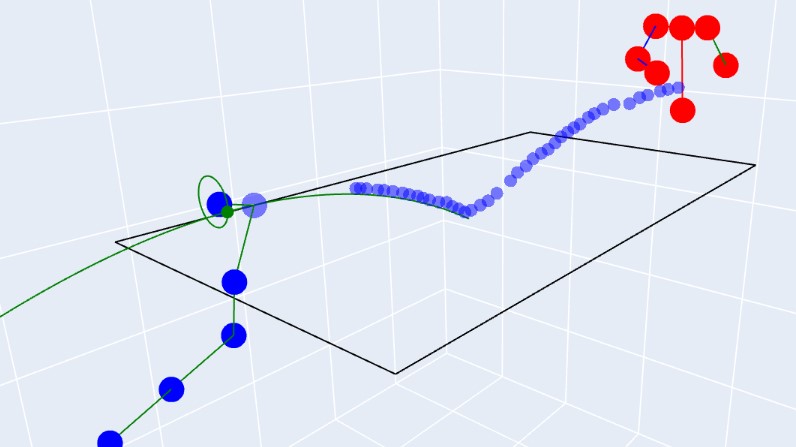}
    \caption{Animation of a frame during simulation. The ball trajectory observed is in blue, and the forecasted trajectory is in green. Skeletons of the KUKA (blue) and opponent (red) in their current positions are also represented.}
    \label{fig:sim}
\end{figure}

\subsection{Visual Servoing Algorithm and Goal State Updates}
We control the robot end effector by estimating the strike point in its trajectory in the $xz$ plane centered at $y$ = -140 with the paddle facing the opponent side. We set the goal state to be the updated strike point, which means the desired velocity vector points from the paddle's current position towards the goal state, and we define an $\alpha$ that represents the proportion of the maximum joint speed at which the robot tracks the desired. This $\alpha$ is always $1.0$ for visual servoing and varies for our different pre-hit inference models. 

We use the same visual servoing model for all experiments, and it always starts at the same timestep after the hit regardless of pre-hit inference models. We implement a simple visual servoing algorithm that estimates the ball trajectory by interpolating 2 quadratic trajectories, a pre-bounce curve and a post-bounce curve, from observed ball positions in real time. For estimating post-bounce curves prior to any observed post-bounce ball positions, we estimate the curve assuming an elastic collision upon bounce. In correspondence with the 0.1 $s$ update frequency of the robot, given our ball is tracked at 100 $hz$, we update the goal position of the ball every 10 frames.

\section{Model Selection, Hyperparameters, and Training}
\label{sec:hyperparameter-sweep}
\subsection{Model and Hyperparameter Selection}
We considered standard Recurrent Neural Networks (RNNs), Long Short-term Memory networks (LSTM), and Gated Recurrent Unit (GRU) networks. We chose these recurrent-based architectures because they are capable of processing variable length time series and can also output intermediate predictions at any given point in the time series. For all three architectures, our model included the recurrent network, followed by an MLP on the final hidden state. Intermediate predictions are computed by applying the trained MLP on intermediate hidden states.

For each architecture, we performed the following grid search over all combinations of hyperparameters, evaluated using 5-fold cross validation. We chose the model that performed best on average across the 5 folds, irrespective of base model architecture. 

\begin{table}[h!]
\centering
\begin{tabular}{|l|l|}
\hline
Parameter & Values \\
\hline
Learning Rate & 1e-02, 1e-03, 1e-04 \\
Hidden Size in Each Recurrent Layer & 100, 200, 400 \\
Hidden Size in MLP & 400, 600, 800, 1200 \\
Percent Dropout & 0.05, 0.1, 0.15 \\
Number of Recurrent Layers & 2, 4, 8 \\
Number of Hidden Layers in MLP & 3, 4, 5 \\
\hline
\end{tabular}
\caption{Hyperparameters in Grid Search}
\label{table:hyperparameters}
\end{table}

\subsection{Training}
We trained all networks using the Adam optimizer with weight decay with a batch size of 500. We used all default AdamW hyperparameters from the built in PyTorch library, except for learning rate, which we included in the hyperparameter sweep. All models were trained for a maximum of 200 epochs, with early stopping if the minimum validation loss did not decrease by more than $0.01$ for $30$ epochs. The GPU used to train was an Nvidia GeForce GTX  $3090$ ($24$GB VRAM). 



\section{Acknowledgements}
We would like to express sincere gratitude to Gaurav Bhatnagar, (Chris) Wei Xun Lai, Andrew Zhang, and Andrew Shen for their contributions to the ball detector, OptiTrack system setup, data segmentation software, and learning experiments, respectively. Their improvements to the data collection and analysis pipeline were invaluable to the successful completion of this research.

\end{appendices}

\end{document}

%% file: preamble.tex
\usepackage[dvipsnames]{xcolor}
\usepackage{amsmath}

\definecolor{fuchsia(web)}{rgb}{0.65, 0.2, 0.65}

